\documentclass[lettersize,journal]{IEEEtran}
\usepackage{amsmath, amssymb, amsfonts}
\usepackage{algorithmic}
\usepackage{algorithm}
\usepackage{array}
\usepackage[caption=false,font=normalsize,labelfont=sf,textfont=sf]{subfig}
\usepackage{textcomp}
\usepackage{stfloats}
\usepackage{url}
\usepackage{verbatim}
\usepackage{graphicx}
\usepackage{cite}
\usepackage{multirow}
\usepackage{multicol} 
\usepackage{stfloats}
\usepackage{hyperref} 
\def\BibTeX{{\rm B\kern-.05em{\sc i\kern-.025em b}\kern-.08em
    T\kern-.1667em\lower.7ex\hbox{E}\kern-.125emX}}
\usepackage{balance}

\begin{document}

\title{DashFusion: Dual-stream Alignment with Hierarchical Bottleneck Fusion for Multimodal Sentiment Analysis}

\author{Yuhua Wen, Qifei Li, Yingying Zhou, Yingming Gao,~\IEEEmembership{Member,~IEEE,} Zhengqi Wen,~\IEEEmembership{Member,~IEEE,} \\ Jianhua Tao,~\IEEEmembership{Senior Member,~IEEE,} and Ya Li,~\IEEEmembership{Member,~IEEE}
\thanks{This work is supported by the National Key R\&D Program of China under Grant No.2024YFB2808802. (Corresponding author: Ya Li.)}
\thanks{Yuhua Wen, Qifei Li, Yingying Zhou, Yingming Gao, and Ya Li are with the School of Artificial Intelligence, Beijing University of Posts and Telecommunications, Beijing 100876, China (e-mail: yuhuawen@bupt.edu.cn; liqifei@bupt.edu.cn; yingyingzhou@bupt.edu.cn; yingming.gao@bupt.edu.cn ; yli01@bupt.edu.cn).}
\thanks{Zhengqi Wen is with the Beijing National Research Center for Information
Science and Technology, Tsinghua University, Beijing 100084, China (e-mail: zqwen@tsinghua.edu.cn).}
\thanks{Jianhua Tao is with the Department of Automation, Tsinghua University, Beijing 100084, China, and also with the Beijing National Research Center for Information Science and Technology, Tsinghua University, Beijing 100084, China (e-mail: jhtao@tsinghua.edu.cn).}}

\markboth{Journal of \LaTeX\ Class Files,~Vol.~14, No.~8, August~2021}%
{Shell \MakeLowercase{\textit{et al.}}: A Sample Article Using IEEEtran.cls for IEEE Journals}


\maketitle
\begin{abstract}
Multimodal sentiment analysis (MSA) integrates various modalities, such as text, image, and audio, to provide a more comprehensive understanding of sentiment. However, effective MSA is challenged by alignment and fusion issues. Alignment requires synchronizing both temporal and semantic information across modalities, while fusion involves integrating these aligned features into a unified representation. Existing methods often address alignment or fusion in isolation, leading to limitations in performance and efficiency. To tackle these issues, we propose a novel framework called Dual-stream Alignment with Hierarchical Bottleneck Fusion (DashFusion).
Firstly, dual-stream alignment module synchronizes multimodal features through temporal and semantic alignment. Temporal alignment employs cross-modal attention to establish frame-level correspondences among multimodal sequences. Semantic alignment ensures consistency across the feature space through contrastive learning. Secondly, supervised contrastive learning leverages label information to refine the modality features. Finally, hierarchical bottleneck fusion progressively integrates multimodal information through compressed bottleneck tokens, which achieves a balance between performance and computational efficiency.
We evaluate DashFusion on three datasets: CMU-MOSI, CMU-MOSEI, and CH-SIMS. Experimental results demonstrate that DashFusion achieves state-of-the-art performance across various metrics, and ablation studies confirm the effectiveness of our alignment and fusion techniques. The codes for our experiments are available at \url{https://github.com/ultramarineX/DashFusion}.
\end{abstract}


\section{Introduction}
\IEEEPARstart{S}{entiment} analysis aims to understand the attitudes and views of opinion holders with computers, becoming a cutting-edge interdisciplinary subject of information technology and psychology \cite{Lu2023SAsurvey}. It can be widely utilized in diverse areas such as recommendation systems \cite{rosa2018knowledge}, health care applications \cite{ghosh2023multimodal}, online education and e-learning \cite{spatiotis2020sentiment}, and mobile assistants \cite{clavel2015sentiment}.

With the rapid development of multimodal learning \cite{baltruvsaitis2018multimodal} and multi-source information fusion \cite{zhao2020multi}, increasing attention has been devoted to inferring and understanding human sentiment using diverse data modalities, such as audio, text, and image. This field is known as multimodal sentiment analysis (MSA). Compared with traditional text-based sentiment analysis, MSA leverages additional modalities to provide a more comprehensive and nuanced understanding of sentiment \cite{gandhi2023multimodal}. However, this expanded scope also introduces significant challenges in effectively utilizing and integrating information from multiple modalities. Alignment and fusion of multimodal data are two primary challenges in MSA.

\begin{figure}[t]
  \centering
  \includegraphics[width=0.95\columnwidth]{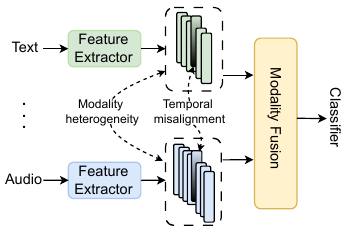}
  \caption{The temporal misalignment and modality heterogeneity in the pipeline of multimodal sentiment analysis (MSA).}
  \label{fig:pipline}
\end{figure}

Alignment refers to the process of ensuring that the information from different modalities is temporally and semantically consistent. 
Let us consider a scenario in which a woman says "I am so happy" while the expression is anger and the speech tone is sorrow. The additional modal information makes it difficult to infer the sentiment of the woman. In fact, MSA typically involves decomposing a video into its constituent components, including textual transcripts, visual frames, and audio signals. Features are then extracted from each modality independently, as shown in Fig. \ref{fig:pipline}. Due to differences in sampling rates and preprocessing methods (i.e., the audio is sampled at 16 kHz, while the video is sampled at 30 frames per second), the features extracted from various modalities at the same timestamp may not correspond accurately, resulting in misalignment in the time dimension. This temporal misalignment makes confused meaning and impairs the model’s ability to accurately infer sentiment.
\IEEEpubidadjcol

Furthermore, misalignment extends beyond the temporal domain into the semantic space due to the diverse qualities, structures and representations across modalities, which is called modality heterogeneity \cite{liang2024foundations}. Each modality operates within distinct representational spaces and possesses unique characteristics, making seamless integration more complex \cite{zhu2024vision+}. Contrastive learning can draw samples closer in semantic space through maximizing the mutual information (MI) between two samples \cite{oord2018representation}. Therefore, researchers have explored semantic alignment using contrastive learning \cite{li2021albef, li2024enhancing}, which has been verified effective in improving model performance. While previous studies have investigated alignment in either the temporal dimension \cite{tsai2019MuIT, guo2022CHFN} or semantic domain \cite{zong2023acformer, li2024enhancing} individually, few works have addressed both simultaneously \cite{LIU2024TSA}, despite their importance in achieving comprehensive alignment.

While multimodal alignment ensures that different modalities are synchronized and consistent, effective multimodal fusion techniques are essential for integrating these aligned modalities into a unified representation that captures the full information.
Various fusion mechanisms have been proposed in recent works to achieve this \cite{zhu2023multimodal}. The two basic and most widely used approaches are: (1) directly applying cross-modal attention between modality features \cite{Wang2024CENet, Cheng2024MMF}, and (2) employing self-attention on concatenated unimodal features \cite{hu-etal-2022-unimse, guo2022dynamically}. However, the former may lead to sub-optimal results as it relies solely on cross-modal attention. The latter results in high computational costs due to the quadratic complexity of the attention mechanism. Additionally, some studies \cite{lv2021PMR, sun2023EMT} have introduced information hubs as intermediaries to enhance communication between modalities, but these approaches often include excessive redundant information, which can negatively affect performance.

Based on these observations, we propose a Dual-stream Alignment with Hierarchical Bottleneck Fusion (DashFusion) framework for MSA, comprising four components: modality encoding, dual-stream alignment, supervised contrastive learning, and hierarchical bottleneck fusion. Firstly, we encode each modality with corresponding feature extractors and encoders. For multimodal data with temporal information, we then perform dual-stream alignment to ensure comprehensive alignment in both the temporal and semantic spaces. For temporal alignment, we establish frame-level correspondences by computing cross-modal attention between text features and other features for aligning non-linguistic modality information to text features at each timestep to obtain well-aligned  multimodal features. For semantic alignment, features from different modalities within the same video are brought closer in the feature space, reducing modality heterogeneity.
We further introduce supervised contrastive learning for both unimodal and multimodal features to enhance feature discrimination and improve model robustness. Drawing inspiration from the information bottleneck, which compresses information by filtering out irrelevant noise while retaining key information \cite{shwartz2017opening, Hu2024IBsurvey}, we extend this concept from feature dimension to sequence length dimension. Hierarchical bottleneck fusion balances performance and computational cost by employing a bottleneck mechanism similar to \cite{nagrani2021IBfusion} and achieves information compression by reducing the number of bottleneck tokens layer by layer. The information bottleneck mechanism improves the generalization ability of our model \cite{kawaguchi2023IB} and the progressive compression strategy forces the model to learn the most beneficial sentiment representation.
To verify the effectiveness of our method, we evaluate DashFusion on three benchmark datasets for MSA, that is, CMU-MOSI, CMU-MOSEI, and CH-SIMS. Experimental results demonstrate that our DashFusion outperforms currently advanced approaches for sentiment analysis. The main contributions of this article can be summarized as follows.

\begin{itemize}
    \item We propose a dual-stream alignment  strategy that ensures temporal and semantic alignment. It not only synchronizes multimodal sequences in temporal dimension, but also enhances the consistency of modality feature in semantic space, leading to a more comprehensive alignment.
    \item We devise a novel multimodal fusion method called hierarchical bottleneck fusion (HBF), which integrates different modality information through bottleneck and removing irrelevant information by compressing bottleneck layer by layer.
    \item Extensive experiments on three publicly available MSA datasets demonstrate that DashFusion achieve the state-of-the-arts (SOTA) performance. Further ablation studies verify the necessity of alignment and validity of our fusion mechanisms.
\end{itemize}

The remainder of this paper is organized as follows: In Section II, we focus on the related works, including multimodal alignment, multimodal fusion, and contrastive learning. In Section III, we introduce the framework of our proposed method in detail. In Section IV, we illustrate the experiments and datasets. In Section V, we present and analyze the experimental results. Finally, we summarize this paper and discuss future work in Section VI.

\section{Related Work}

\subsection{Multimodal Alignment}

Multimodal alignment is to identify the direct relations between (sub)elements from two or more different modalities. According to the purpose of alignment, mainstream alignment approaches can be categorized into two types: temporal alignment and semantic alignment.
 
As mentioned in the section above, temporal alignment makes tokens from different modality features at the same timestamp correspond correctly. Alignment of text, speech and video in the time dimension has long been a key issue in video understanding. Early works \cite{zadeh2016mosi, zadeh2018mosei} utilized auxiliary software such as P2FA for word-level alignment, which is very time-consuming. Because cross-modal attention (CA) can model the global dependencies between two sequences, recent works employed CA to learn the interactions between multimodal sequences across different time steps and to adaptively align streams from one modality to another. Tsai et al. \cite{tsai2019MuIT} proposed Cross-Modal Transformers, which learn cross-modal attention to enhance the target modality. Guo et al. \cite{guo2022CHFN} dynamically adjusted word representations in different non-verbal contexts using unaligned multimodal sequences.

Semantic alignment learns how to latently align the paired data in semantic space, which is often used as an intermediate step for downstream tasks. Since the core idea of contrastive learning aligns with semantic alignment, many works achieve semantic alignment by contrastive learning. ALBEF \cite{li2021albef} introduced a contrastive loss to align the image and text representations before fusing them through cross-modal attention, which enables more grounded vision and language representation learning. Zong et al. \cite{zong2023acformer} proposed ACFormer, which explicitly aligns different modality features by contrastive learning before pivot attention fusion. Li et al. \cite{li2024enhancing} aligned emotional information in audio-video representations through contrastive learning to enhance the effectiveness of later modality fusion.

Inspired by their success, we design a dual-stream alignment to align text, audio, and video in both time and semantic dimensions. It should be noted that our dual-stream alignment not only adds two alignment methods simply but also forms the basis of further supervised contrastive learning and hierarchical bottleneck fusion.

\subsection{Multimodal Fusion}

Multimodal fusion integrates information from multiple modalities with the goal of predicting an outcome measure. In MSA, fusion methods primarily focus on designing sophisticated fusion mechanisms to obtain joint representations of multimodal data \cite{zhu2023multimodal}.
According to the fusion method, previous works can be classified into two categories: early fusion and late fusion.

Early fusion merges the features of each modality extracted at the input level to construct a joint representation. For example, Zadeh et al. \cite{zadeh2017TFN} used Tensor Fusion Networks (TFN) to obtain a tensor representation by computing the outer product of unimodal representations. Liu et al. \cite{liu2018LMF} designed a low-rank multimodal fusion (LMF) method to reduce the computational complexity of tensor-based approaches. Han et al. \cite{han2021MMIM} proposed MMIM, which improves multimodal fusion through hierarchical mutual information maximization.

Late fusion first learns sentiment information based on each modality, and then proposes different mechanisms to incorporate each modality features into the final decision.
Lv et al. \cite{lv2021PMR} introduced a message center to explore tri-modal interactions and perform progressive multimodal fusion.
These methods perform fusion directly without considering the misalignment between the different modality features, which will lead to suboptimal results. Yang et al. \cite{yang2022FDMER} employed adversarial learning to decouple modality features and used cross-modal attention for modality fusion. Li et al. \cite{li2023DMD} proposed a decoupled multimodal distillation (DMD) approach that facilitates flexible and adaptive cross-modal knowledge distillation to enhance the modality features and then fuse them.

Although these methods have achieved good performance, they fail to consider the impact of redundant information and fully exploit complementary information, which limits their performance in MSA. Therefore, we use bottleneck as intermediaries to enhance communication between modalities and design a hierarchical mechanism to discard redundant information.

\subsection{Contrastive Learning}

Contrastive learning learns better data representation by pulling similar samples closer together while pushing dissimilar samples further apart in the feature space. Since it does not require labels, contrastive learning has achieved significant success in self-supervised learning \cite{chen2020simclr,he2020MOCO}. Khosla et al. \cite{khosla2020supcon} extended contrastive learning to the supervised setting. They contrast samples by different classes and find it more stable for hyperparameters. 

Given that multimodal data naturally contains inherent positive and negative pair relationships, contrastive learning has been widely applied in multimodal learning. CLIP \cite{radford2021clip} achieved remarkable performance on image-text tasks by using a contrastive loss and a huge dataset. Sun et al. \cite{sun2022MCL4video} used contrastive learning to learn temporal relationships for video-language models. Basil et al. \cite{mustafa2022MCL4MOE} trained a mixture of experts model capable of multimodal learning by contrastive learning. Jiang et al. \cite{Jiang2024CL4MLLM} adapted contrastive learning to multimodal large language models (MLLMs) to reduce hallucinations.

Recently, some MSA methods obtain modality representations based on contrastive learning. HyCon \cite{mai2022HyCon} simultaneously performed intra-/inter-modal contrastive learning to obtain tri-modal joint representations. Yang et al.\cite{yang2023ConFEDE} decomposed each modality feature into similar and dissimilar parts for text-centered contrastive learning and designed a data sampler to retrieve positive/negative pairs. Fan et al.\cite{Fan2024MCL} designed multi-level contrastive learning to alleviate the heterogeneity in MSA.
However, the existence of modality gap \cite{liang2022modalitygap} makes it difficult to use contrastive learning alone to capture complementary information across different modalities. Therefore, we decide to not only leverage contrastive learning for modality alignment, but also use sentiment label information to refine modality features by supervised contrastive learning for further multimodal fusion.

\begin{figure*}[ht]
  \centering
  \includegraphics[width=0.95\textwidth]{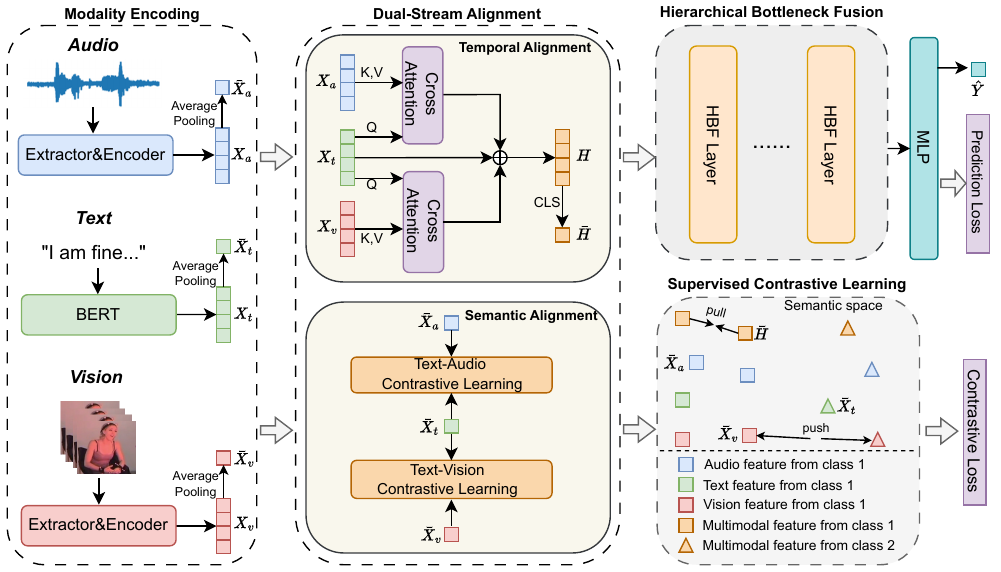}
  \caption{The overall architecture of the DashFusion model for MSA. It consists of modality encoding, dual-stream alignment, supervised contrastive learning, and hierarchical bottleneck fusion. Features are initially encoded independently, aligned temporally and semantically, refined through supervised contrastive learning, and finally fused via hierarchical bottleneck layers to produce robust sentiment predictions.}
  \label{fig:framework}
\end{figure*}

\section{Methodology}

The overall architecture of DashFusion is illustrated in Fig. \ref{fig:framework}. It consists of four parts: modality encoding, dual-stream alignment, supervised contrastive learning, and hierarchical bottleneck fusion. Our model first encodes each modality with corresponding feature extractors and encoders. Then, unimodal features are fed into the dual-stream alignment module to align in both the time dimension and feature space, producing aligned multimodal features. After that, the supervised contrastive learning module is employed to enhance the model's ability to distinguish different sentiments. Finally, we apply hierarchical fusion of modal features using the concept of information bottleneck. The fused unimodal features and multimodal features are concatenated and used to predict the sentiment score. Below, we present the details of the four parts of DashFusion.

\subsection{Preliminary}

The Transformer model effectively captures global dependencies in sequential data using scaled dot-product attention. Each Transformer layer comprises self-attention (SA), layer normalization (LayerNorm) \cite{ba2016layernorm}, and a feed-forward network (FFN) with residual connection \cite{he2016res}. The computations of a Transformer layer can be represented as follows:

\begin{equation}
  \label{eq}
  \begin{aligned}
    \operatorname{Transformer-layer}(X)&=\operatorname{Layernorm}(Z+FFN(Z))  \\
    Z&=\operatorname{LayerNorm}(X+SA(X))
  \end{aligned}
\end{equation}

where SA denotes self-attention, which is the core component in Transformer. For the input sequence X, we define the Query as $Q=XW_Q$, Key as $K=XW_K$, Value as $V=XW_V$, where $W_Q$, $W_K \in \mathbb{R}^{d \times d_k}$, $W_V \in \mathbb{R}^{d \times d_v}$ are learnable parameters and $d_k$ is the dimension of the attention head. The SA can be formulated as follows:

\begin{equation}
  \label{eq3}
  \begin{aligned}
    \operatorname{SA}(X) &=\operatorname{softmax}\left(\frac{Q K^{T}}{\sqrt{d_{k}}}\right)V \\
    &= \operatorname{softmax}\left(\frac{XW_Q W_K^{T}X^{T}}{\sqrt{d_{k}}}\right)XW_V 
  \end{aligned}
\end{equation} 

Cross-modal attention (CA) involves two modalities. The Query is from the target modality $t$, while the Key and Value are from the source modality $s$. The Query as $Q=X_tW_Q$, Key as $K=X_sW_K$, Value as $V=X_sW_V$. In this way, CA can provide a latent adaptation from modality $s$ to $t$ :

\begin{equation}
  \label{eq3}
  \begin{aligned}
    \operatorname{CA}(X_t,X_s) &=\operatorname{softmax}\left(\frac{Q_{t} K_{s}^{T}}{\sqrt{d_{k}}}\right)V_s \\
    &= \operatorname{softmax}\left(\frac{X_tW_Q W_K^{T}X_s^{T}}{\sqrt{d_{k}}}\right)X_sW_V 
  \end{aligned}
\end{equation}

Note that, for simplicity, we only present the formulation of single-head attention. In practice, we use multi-head CA (MHCA) to allow the model to attend to information from different feature subspaces.

\subsection{Modality Encoding}

Regarding the multimodal inputs, which consist of video, audio, and text data, these raw inputs are initially transformed into feature vectors for subsequent processing. Following the approach of previous works \cite{yu2021Self-MM, han2021MMIM}, both the audio and visual inputs are first converted into numerical sequential vectors through feature extractors. These extractors are  non-trainable components and are applied to extract low-level features from the raw data.
Once these raw multimodal inputs are converted into their corresponding feature vectors, we employ two separate transformer encoders to process the sequential features for the visual and audio modalities. These encoders capture the temporal dependencies and contextual relationships within each modality sequence. For the text modality, we utilize the pre-trained BERT model to encode the text data. The output feature representation from BERT is then scaled to match the feature dimension of the other modalities, ensuring consistency in the feature space across all input types.

The resulting modality-specific features are denoted as $X_m \in \mathbb{R}^{T_m \times d_m} $, where $m \in \{t, v, a\}$, $T_m$ is the sequence length and $d_m$ is the vector dimension of each modality. In practice, both $T_m$ and $d_m$ may vary depending on the dataset used.

\subsection{Dual-Stream Alignment}

We propose a dual-stream alignment method that includes both temporal and semantic alignment for comprehensive alignment. For temporal alignment, the unimodal features are dynamically aligned in the time dimension. For semantic alignment, we align matching modal pairs in the feature space. Furthermore, we choose text features as center in both temporal and semantic alignment, which can be viewed as connecting temporal and semantic alignment through text.


\subsubsection{Temporal Alignment}

Inspired by previous works such as \cite{tsai2019MuIT, guo2022CHFN}, which utilize attention mechanisms to achieve dynamic alignment, we incorporate cross-modal attention (CA) to handle temporal alignment across different modalities. Temporal alignment refers to the process of aligning tokens from various modality sequences based on their semantic content, ensuring that features from distinct modalities correspond effectively at each time step. Cross-modal attention is particularly effective in modeling global dependencies between two modality sequences, as it captures interactions across the entire sequence rather than relying solely on local temporal information.

Compared to continuous and high-dimensional visual and audio signals, text data is discrete and contains more explicit semantic information. This explicitness makes text an ideal modality for use as a benchmark during alignment, as it allows for precise, word-level correspondence. As such, we align the visual and audio features to the text modality, using text as the reference for alignment. 
Firstly, we employ a cross-modal attention mechanism to establish correspondences between modalities. Specifically, we select text features $X_t$ as the Query, and the audio features  $X_a$ act as the Key and Value to compute the cross-modal representation $X_{a \to v}$, which transfers information from audio to text. Similarly, text features $X_t$ are used as the Query, with visual features  $X_v$ acting as the Key and Value to derive $X_{v \to t}$. Once the cross-modal representations are obtained, they are added to the original text features $X_t$, resulting in the final well-aligned multimodal representation $H$, which can be formalized as:

\begin{equation}
  \begin{aligned}
    H &= X_t + X_{a \to t} + X_{v \to t} \\
    &= X_t + \operatorname{CA}(X_t,X_a)+\operatorname{CA}(X_t,X_v)
  \end{aligned}
\end{equation}

\subsubsection{Semantic Alignment}

Semantic alignment aims to draw close the features of matching modal pairs in feature space. Images, text, and audio from the same video are considered matching modal pairs. To achieve this, we utilize contrastive learning for semantic alignment. This process maximizes a lower bound on the mutual information (MI) between different "views" of a video.

Notably, we select text as the sole anchor for semantic alignment, meaning the modality pairs in contrastive learning are text-audio and text-vision. This design choice is motivated by two key reasons. First, multimodal sentiment analysis remains predominantly text-centric, with visual and audio modalities exhibiting substantially higher noise levels compared to text \cite{Jin2023MSEN}. Second, this approach maintains consistency with temporal alignment principles, as contrastive learning can be conceptually framed as a dictionary lookup operation where the text anchor serves as the query.

Specifically, we employ the NT-Xent loss \cite{chen2020simclr} as the loss function for contrastive learning. The loss for sample $i$ is defined as follows.
\begin{equation}
  \label{eq1}
  \ell_{\mathrm{cl}}^{i}=\sum_{(a, p) \in \mathcal{P}_{i}}-\log \frac{\exp (\operatorname{sim}(a, p) / \tau)}{\sum_{(a, k) \in \mathcal{N}_{i} \cup \mathcal{P}_{i}} \exp (\operatorname{sim}(a, k) / \tau)}
\end{equation}
where $\tau$ is a temperature hyperparameter \cite{wu2018unsupervised} that controls the scaling of similarity scores in the contrastive loss. A lower $\tau$ sharpens the distinction between positive and negative pairs by amplifying differences in similarity, while a higher $\tau$ smooths the distribution, making the model less sensitive to these differences. Proper tuning of $\tau$ is crucial for optimizing the model's performance. $(a,p)$, $(a,k)$ correspond to the global features $\bar{X}_m$ of each modality, which are obtained by average pooling the unimodal features $X_m$ along the time dimension.
$a$ represents the anchor in contrastive learning. $\mathcal{P}$ is the set of positive samples, and $\mathcal{N}$ is the set of negative samples. The similarity is measured by the dot product of the encoded anchors and a set of encoded samples.

\subsection{Supervised Contrastive Learning}

To enhance the robustness of DashFusion and fully utilize the information provided by the labels, we introduce supervised contrastive learning and unify it with semantic alignment in an NT-Xent loss framework. To maximize the potential of contrastive learning, we employ the hard negative mining approach. We construct positive and negative sample sets similar to the data sampler in ConFEDE \cite{yang2023ConFEDE}, which retrieves similar samples for a given sample based on both features and labels across samples. Different from other methods, not only unimodal features $X_m$ for supervised contrastive learning but also multimodal features $H$ obtained by temporal alignment are selected in supervised contrastive learning.

In practice, we first compute the cosine similarity between each sample pair $(i,j)$ in the dataset $D$, and then retrieve sets of similar and dissimilar samples for each anchor sample $i$. For positive pairs, we randomly select two samples from the same class with high cosine similarity. For negative pairs, we select four samples with different labels: two of them are similar to sample $i$ and the other two are dissimilar. This ensures a robust contrastive learning process that improves the alignment and generalization of multimodal features.

\subsection{Hierarchical Bottleneck Fusion}

\begin{figure*}[t]
  \centering
  \includegraphics[width=0.95\textwidth]{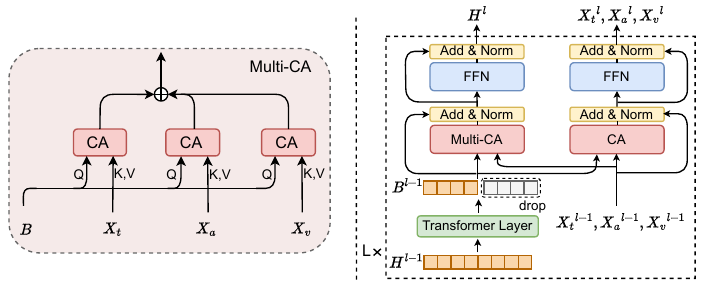}
  \caption{Hierarchical Bottleneck Fusion (HBF) layer architecture (right). Multi-CA (left) gathers and integrate information from different modality features through cross-modal attention (CA).}
  \label{fig:HBF layer}
\end{figure*}

Although the multimodal feature $H$ integrates information from audio, visual, and text modalities through temporal alignment, the interaction is unidirectional (from text to vision and audio) and insufficient, limiting the performance of $H$ in downstream tasks. To enable richer cross-modal interactions across multiple layers, we introduce a bottleneck as a communication hub to facilitate bidirectional information flow among three modalities. The number of bottleneck tokens progressively reduces at each layer, compressing the multimodal information while filtering out irrelevant details. This progressive design allows the model to effectively integrate and compress multimodal information.

The structure of the Hierarchical Bottleneck Fusion (HBF) layer is illustrated in Fig. \ref{fig:HBF layer}. Suppose the Hierarchical Bottleneck Fusion (HBF) module consists of $L$ layers. Initially, the bottleneck representation $B^{l}$ for the $l$-th layer is derived by applying a Transformer encoder to the multimodal feature $H^{l-1}$ and selecting the first $p/2^{l-1}$ tokens.
\begin{equation}
    B^{l} = \operatorname{Transformer-layer}(H^{l-1})[0:p/2^{l-1}]
\end{equation}
where $B^{l}$ is the bottleneck of the $l$-th layer, $H^{l-1}$ is the multimodal feature from the previous layer.
The Transformer layer serves to redistribute the multimodal information, ensuring that the most relevant content is captured in these tokens. The bottleneck acts as a compact summary of the multimodal input, serving as an intermediary to efficiently fuse information.

Next, multimodal fusion in each layer proceeds in two stages. First, the bottleneck tokens $B^{l}$ act as the Query to compute cross-modal attention with each of the three unimodal features (text, image, and audio), compressing the information from various modalities into the bottleneck.
\begin{equation}
  \begin{aligned}
    H^{l} &= \operatorname{LayerNorm}\left(B^{l}+\operatorname{Multi-CA}(B^{l},X_{m}^{l-1})\right) \\
    H^{l} &=\operatorname{LayerNorm}\left(H^{l}+\operatorname{FFN}\left(H^{l}\right)\right)
  \end{aligned}
\end{equation}
where $H^{l}$ is multimodal features at the $l$-th layer. $X_{m}^{l-1}, m \in \{t, v, a\}$ represents the unimodal features in $l-1$-th layer. LayerNorm denotes layer normalization \cite{ba2016layernorm}, FFN is a feed-forward network with two linear transformations and a ReLU activation \cite{glorot2011deep}. Multi-CA refers to multi cross-modal attention, where information is aggregated from all unimodal sources. It is formalized as:
\begin{equation}
  \begin{aligned}
    &\operatorname{Multi-CA}(B^{l},X_{m}^{l-1})=B^{l}+CA(B^{l},X_{m}^{l-1})
  \end{aligned}
\end{equation}

Then, the unimodal features at the $l$-th layer $\mathbf{X}_{m}^{l}$ are updated by cross-modal attention with the refined bottleneck representation $B^{l}$.
\begin{equation}
  \begin{aligned}
    Z_{m}^{l}&=\operatorname{LayerNorm}\left(X_{m}^{l-1}+\operatorname{CA}({X}_{m}^{l-1},B^{l}) \right) \\
    X_{m}^{l}&=\operatorname{LayerNorm}\left(Z_{m}^{l}+\operatorname{FFN}\left(Z_{m}^{l}\right)\right),m \in \{t, v, a\}
  \end{aligned}
\end{equation} 

This process allows each modality to incorporate information from others. The number of bottleneck tokens is halved in each subsequent layer, ensuring progressive information compression while retaining the essential features for accurate sentiment analysis.

\subsection{Overall Learning Objectives}

The DashFusion model is trained with a multitask learning objective function, which consists of prediction loss and contrastive loss.

\textbf{Prediction Loss.} 
A multilayer perceptron (MLP) with the ReLU activation function is used as a classifier to obtain the final prediction. We concatenate the first token of unimodal features and the bottleneck features after fusion to obtain the inputs to the classifier. The prediction loss is calculated by mean squared error.

\begin{equation}
    \mathcal{L}_{pred}=\frac{1}{n} \sum_{i=1}^{N}(y_{i}-\hat{y}_{i})^{2} \\
\end{equation}
where $n$ is the number of training samples and $y_i$ is the sentiment label.

\textbf{Contrastive Loss.}
As mentioned above, we unify the two modules of semantic alignment and supervised contrastive learning through a simple joint contrastive loss. Specifically, this contrastive loss is expressed as:

\begin{equation}
    \mathcal{L}_{con}=\frac{1}{n} \sum_{i=1}^{N} \ell_{\mathrm{cl}}^{i} \\
\end{equation}
where $\ell_{\mathrm{cl}}^{i}$ is the contrastive loss of sample $i$.

Finally, the overall loss function of DashFusion is represented as Equation \ref{eq12}, where $ \lambda$ is a hyperparameter to balance the contribution of each component to the overall loss.

\begin{equation}\label{eq12}
    \mathcal{L}_{all} = \mathcal{L}_{pred} + \lambda \mathcal{L}_{con}
\end{equation}

\section{EXPERIMENTAL DATABASES AND SETUP}

\subsection{Datasets}

To evaluate the performance of DashFusion, we conduct experiments on three publicly available datasets in MSA research, CMU-MOSI \cite{zadeh2016mosi}, CMU-MOSEI \cite{zadeh2018mosei}, and CH-SIMS \cite{yu2020ch-sims}. The split specifications of the three datasets are shown in Table \ref{dataset}. Here we give a brief introduction to the above datasets.

\textbf{CMU-MOSI.} As one of the most popular benchmark datasets for MSA, CMU-MOSI contains 2,199 utterance-video clips sliced from 93 videos in which 89 distinct narrators are sharing opinions on interesting topics. Each clip is manually annotated with a sentiment value ranging from -3 (strongly negative) to +3 (strongly positive).

\textbf{CMU-MOSEI.} The dataset comprises 22,856 annotated video clips collected from YouTube. The CMU-MOSEI dataset upgrades CMU-MOSI by expanding the number of samples, utterances, speakers, and topics. Following the annotation method in CMU-MOSI, each utterance is labeled ranging from [-3, +3] to reflect the sentiment intensity.

\textbf{CH-SIMS.} The CH-SIMS dataset is a distinctive Chinese MSA dataset that contains 2,281 refined video clips collected from different movies, TV serials, and variety shows. Each sample has one multimodal label and three unimodal labels with a sentiment score from -1 (strongly negative) to 1 (strongly positive).

\begin{table}[!t]
  \begin{center}
  \caption{Dataset Statistics in CMU-MOSI, CMU-MOSEI, and CH-SIMS.}
  \label{dataset}
  \begin{tabular}{ccccc}
    \hline
    Dataset &Train & Valid & Test & All  \\
    \hline
    CMU-MOSI & 1284 & 229 & 686 & 2199 \\
    CMU-MOSEI & 16326 & 1871 & 4659 & 22856 \\
    CH-SIMS & 1368 & 456 & 457 & 2281 \\
    \hline
  \end{tabular}
  \end{center}
\end{table}

\subsection{Evaluation Metrics}

Following previous works \cite{yu2021Self-MM,han2021MMIM,yang2023ConFEDE}, we report our results for classification and regression with the average of five runs of different seeds. For classification, we report the multi-class accuracy and weighted F1 score, i.e., 2-class accuracy (Acc-2), 3-class accuracy (Acc-3), and 5-class accuracy (Acc-5) for CH-SIMS or 7-class accuracy (Acc-7) for CMU-MOSI and CMU-MOSEI. Moreover, agreeing with prior works \cite{han2021MMIM,yu2021Self-MM,yang2023ConFEDE}, Acc-2 and F1 score on CMU-MOSI and CMU-MOSEI have two forms: negative/non-negative (non-exclude zero) and negative/positive (exclude zero). For regression, we report Mean Absolute Error (MAE) and Pearson correlation (Corr). Except for MAE, higher values indicate better performance for all metrics.

\subsection{Feature Extraction} 

To ensure a fair comparison, we leverage the official unaligned features provided by each benchmark dataset, consistent with the top-performing MSA methods.

\textbf{Text Modality}: Transformer-based pre-trained language models have achieved state-of-the-art results across various natural language processing (NLP) tasks. Following recent works \cite{han2021MMIM,yu2021Self-MM}, we use pre-trained BERT models to encode raw text. Specifically, "bert-base-chinese" \footnote{\url{https://huggingface.co/bert-base-chinese/}} is employed for CH-SIMS, while "bert-base-uncased" \footnote{\url{https://huggingface.co/bert-base-uncased/}} is used for CMU-MOSI and CMU-MOSEI.

\textbf{Audio Modality}: For CMU-MOSI and CMU-MOSEI, we use the COVAREP framework \cite{2014COVAREP} to extract low-level audio features, including pitch, voiced/unvoiced segments, glottal source parameters, and 12 Mel-frequency cepstral coefficients (MFCCs), among others. For CH-SIMS, we employ Librosa \cite{mcfee2015librosa} to obtain audio representations, extracting features such as the logarithmic fundamental frequency (log F0), 20 MFCCs, and 12 Constant-Q Transform (CQT) chromagrams.

\textbf{Vision Modality}: For CMU-MOSI and CMU-MOSEI, Facet \footnote{\url{https://imotions.com/platform/}} is utilized to extract 35 facial action units, capturing facial muscle movements linked to emotions. In contrast, for CH-SIMS, we use the OpenFace2.0 toolkit \cite{OpenFace} to obtain visual features, including facial action units, facial landmarks, and head pose. These visual features are sampled at 30 Hz, forming a sequence that captures facial gestures over time.

\subsection{Implementation Details}

\subsubsection{Model Details}

For modality encoder, we use transformers with 128 dimensions as Audio and Vision Encoders. For CH-SIMS and CMU-MOSI, we use two single-layer transformer encoders, while for CMU-MOSEI, we use three transformer layers due to its larger dataset size. In hierarchical bottleneck fusion (HBF), we set the number of bottleneck tokens \( p \) to 8. The number of fusion layers is set to 2 for CMU-MOSI and CH-SIMS, and 3 for CMU-MOSEI.

\subsubsection{Training Details}

All experiments we conduct on a single NVIDIA RTX 4090 GPU, with DashFusion comprising fewer than 120 million parameters across all implementations. We train DashFusion for MSA using the Adam optimizer \cite{kingma2014adam} with the aforementioned encoders. For CH-SIMS and CMU-MOSI, we train DashFusion for 100 epochs with a learning rate of 5e-5 and a batch size of 16. For CMU-MOSEI, we train the model for 25 epochs with a batch size of 8 and a learning rate of 2e-5. The loss ratio \( \lambda \) is set to 0.2 and the temperature in NT-Xent loss is set to 0.5.

\subsection{Baselines}

To comprehensively validate the performance of our model, we compare our method with several advanced and state-of-the-art baselines in Table \ref{exp:mosi} and \ref{exp:sims}: TFN \cite{zadeh2017TFN}, LMF \cite{liu2018LMF}, MulT \cite{tsai2019MuIT}, MAG-BERT \cite{rahman2020MAG-BERT}, MISA \cite{hazarika2020MISA}, Self-MM \cite{yu2021Self-MM}, MMIM \cite{han2021MMIM}, ConFEDE \cite{yang2023ConFEDE}. To ensure fairness in comparison, methods that only report results from a single run and lack valid official code for reproduction are not selected. Additionally, methods that utilize pretrained audio or vision models are not considered.

\textbf{TFN.} The Tensor Fusion Network \cite{zadeh2017TFN} calculates a multi-dimensional tensor utilizing outer product operations to capture uni-, bi-, and tri-modal interactions.

\textbf{LMF.} The Low-rank Multimodal Fusion (LMF) \cite{liu2018LMF} decomposes stacked high-order tensors into many low-rank factors to perform multimodal fusion efficiently.

\textbf{MulT.} The Multimodal Transformer (MulT) \cite{tsai2019MuIT} employs directional pairwise cross-modal attention to capture the interactions among multimodal sequences and adaptively align streams between different modalities.

\textbf{MAG-BERT.} The Multimodal Adaptation Gate for BERT (MAG-BERT) \cite{rahman2020MAG-BERT} designs an alignment gate and inserts it into different layers of the BERT backbone to refine the fusion process.

\textbf{MISA.} The Modality-Invariant and -Specific Representations (MISA) \cite{hazarika2020MISA} projects each modality into modality-invariant and modality-specific spaces with special limitations. Fusion is then accomplished on these features.

\textbf{SELF-MM.} SELF-MM \cite{yu2021Self-MM} assigns each modality a unimodal training task to obtain labels, then jointly learns the multimodal and unimodal representations using multimodal and generated unimodal labels.

\textbf{MMIM.} MMIM \cite{han2021MMIM} proposes a hierarchical MI maximization framework that occurs at the input level and fusion level to reduce the loss of valuable task-related information.


\textbf{ConFEDE.} ConFEDE \cite{yang2023ConFEDE} is based on contrastive feature decomposition, which utilizes a unified contrastive training loss to capture the consistency and differences across modalities and samples.

\begin{table*}[t]
  \centering
  \caption{Comparison on CMU-MOSI and CMU-MOSEI. $\dagger$ Results from \cite{yang2023ConFEDE}, $\ddagger$ Results from \cite{han2021MMIM}. All Other Results are Reproduced Using Publicly Available Source Codes and Original Hyperparameters under the Same Setting. In Acc-2 and F1, the Left of the "/" Corresponds to "Negative/Non-negative" and the Right Corresponds to "Negative/Positive". (A) Means the Model Utilized the Aligned Data.}
  \label{exp:mosi}
  \resizebox{\textwidth}{!}{
    \begin{tabular}{lcccccccccc}
    \hline
    \multirow{2}*{Method} & \multicolumn{5}{c}{CMU-MOSI}  & \multicolumn{5}{c}{CMU-MOSEI} \\
    \cline{2-11} 
    & Acc-2($\uparrow$) & F1($\uparrow$) & Acc-7($\uparrow$) & MAE($\downarrow$) & Corr($\uparrow$)& Acc-2($\uparrow$) & F1($\uparrow$) & Acc-7($\uparrow$) & MAE($\downarrow$) & Corr($\uparrow$)\\
    \hline
    TFN$^{\dagger}$ & -/80.8& -/80.7& 34.9& 0.901& 0.698& -/82.5& -/82.1& 50.2& 0.593& 0.700\\
    LMF$^{\dagger}$ & -/82.5& -/82.4& 33.2& 0.917& 0.695& -/82.0& -/82.1& 48.0& 0.623& 0.677\\
    MuIT(A)$^{\dagger}$ & -/83.0& -/82.8& 40.0& 0.871& 0.698& 81.15/84.63& 81.56/84.52& 52.84& 0.559& 0.733\\
    MISA(A)$^{\dagger}$ & 81.8/83.4& 81.7/83.6& 42.3& 0.783& 0.761& 83.6/85.5& 83.8/85.3& 52.2& 0.555& 0.756\\
    MAG-BERT$^{\dagger}$ & 82.13/83.54& 81.12/83.58& 41.43& 0.790& 0.766& 79.86/83.86& 80.47/83.88& 50.41& 0.583& 0.741\\
    Self-MM$^{\dagger}$ & 83.44/85.46& 83.36/85.43& 46.67& 0.708& 0.796& \textbf{83.76}/85.15& \textbf{83.82}/84.90& \textbf{53.87}& 0.531& 0.765\\
    ConFEDE$^{\dagger}$ & 84.17/85.52& 84.13/85.52& 42.27& 0.742& 0.784& 81.65/85.82& 82.17/85.83& 54.86& 0.522& 0.780\\
    MMIM$^{\ddagger}$ & 84.14/86.06& 84.0/85.98& 46.65& 0.70& 0.800& 82.24/85.97& 82.66/85.94& 54.24& 0.526& 0.772\\
    \hline
    ConFEDE & 82.8/84.76& 82.72/84.74& 41.55& 0.757& 0.775& 81.65/84.53& 81.98/84.36& 52.16& 0.564& 0.746\\
    MMIM & 83.46/85.11& 83.4/85.24& \textbf{46.2}& 0.714& 0.794& 81.64/85.24& 81.84/85.19& 53.23& 0.538& 0.763\\
    Ours & \textbf{84.26/85.82}& \textbf{84.17/85.78}& 45.63& \textbf{0.709}& \textbf{0.796} & 82.27/\textbf{86.3}& 82.7/\textbf{86.24}& 53.12& \textbf{0.524}& \textbf{0.784}\\
    \hline
    \end{tabular} 
  }
\end{table*}

\section{Results and Discussion}

We first conduct comparative experiments between DashFusion and currently advanced multimodal sentiment analysis models. Then we investigate the choice of the number of bottleneck tokens. Subsequently, we reveal the necessity of each component in DashFusion, including dual-stream alignment, supervised contrastive learning, and hierarchical bottleneck fusion. To further verify the effectiveness of our proposed method, we also compare hierarchical bottleneck fusion with other fusion mechanisms for MSA.

\subsection{Performance Comparison}

The performance comparison of all methods on CMU-MOSI, CMU-MOSEI, and CH-SIMS is summarized in Table \ref{exp:mosi} and Table \ref{exp:sims}. The scores of the proposed method and its variations are the averages of 5 runs. The performances of all other baselines have been sourced from published papers or official repositories.

As shown in Table \ref{exp:mosi}, our method yields better or comparable results to many baseline methods, demonstrating the effectiveness of our approach in multimodal sentiment analysis (MSA). Specifically, on CMU-MOSI, our model outperforms all baselines except for Acc-7, where it remains comparable to the best-performing methods. On CMU-MOSEI, our model achieves the highest scores in the NP setting for Acc-2, F1, MAE, and Corr metrics. In particular, we have significant improvements in NP Acc-2 and F1, indicating its superior ability to distinguish between positive and negative sentiments. While Acc-7 lags slightly behind Self-MM and MMIM, this may be due to our method’s less explicit optimization for fine-grained sentiment distinctions. Overall, our approach consistently delivers strong performance across various datasets and evaluation metrics. 

In the negative/non-negative (NN) setting, our method does not perform as well as it does in the negative/positive (NP) setting. This is because the NN setting introduces greater ambiguity, as it requires distinguishing strictly negative samples from both neutral and positive ones. In contrast, the NP setting benefits from a clearer sentiment contrast, making classification easier. Additionally, samples with regression labels near zero often have weak sentiment cues, further challenging the model’s ability to differentiate them accurately.

To further assess the effectiveness of our proposed method, DashFusion, we conduct training on the CH-SIMS dataset. The scenarios in CH-SIMS are more intricate compared to those in CMU-MOSI and CMU-MOSEI, posing a greater challenge for modeling multimodal data. As seen in Table \ref{exp:sims}, for baselines that only use multimodal labels, our method outperforms all of them on all metrics. Compared with the best baseline model, we achieve superior performance on multi-class, outperforming it by 2.14\% on Acc-3 and 2.55\% on Acc-5. Furthermore, our method performs closely to ConFEDE, which uses unimodal labels to enhance model training. Given that unimodal labels are difficult and time-consuming to obtain in real-world scenarios, our method demonstrates a significant advantage. These results highlight the robustness and practical applicability of DashFusion in diverse and complex multimodal sentiment analysis tasks.

\begin{table*}
  \centering
  \caption{Comparison Results on CH-SIMS. $\dagger$ Results from \cite{mao2022m-sena} and its Corresponding GitHub Page \protect\footnotemark[1]. $\ddagger$ Results from \cite{yang2023ConFEDE}. All Other Results are Reproduced Using Publicly Available Source Codes and Original Hyperparameters under the Same Setting. (U) Means the Model Used Multimodal Labels and Unimodal Labels.}
  \label{exp:sims}
  \resizebox{0.6\textwidth}{!}{
  \begin{tabular}{lccccccc}
    \hline
    \multirow{2}*{Method} & \multicolumn{5}{c}{CH-SIMS}  \\
    \cline{2-7} 
    & Acc-2($\uparrow$) & F1($\uparrow$) & Acc-3($\uparrow$) & Acc-5($\uparrow$) & MAE($\downarrow$) & Corr($\uparrow$)\\
    \hline
    TFN$^{\dagger}$ & 78.38 & 78.62 & 65.12 & 39.30 & 0.432 & 0.591 \\
    LMF$^{\dagger}$ & 77.77 & 77.88 & 64.48 & 40.53 & 0.441 & 0.576 \\
    MulT$^{\dagger}$ & 78.56 & 79.66 & 64.77 & 37.94 & 0.453 & 0.561 \\
    MISA$^{\dagger}$ & 76.54 & 76.59 & - & - & 0.447 & 0.563 \\
    MAG-BERT$^{\dagger}$ & 74.44 & 71.75 & - & - & 0.492 & 0.399 \\
    self-MM$^{\dagger}$ & 80.04 & 80.44 & 65.47 & 41.53 & 0.425 & 0.595 \\
    ConFEDE(U)$^{\ddagger}$ & 82.23 & 82.08 & 70.15 & 46.30 & 0.392 & 0.637 \\
    \hline
    MulT & 78.24 & 78.54 & 64.38 & 40.06 & 0.446 & 0.579 \\
    Self-MM & 78.56 & 78.60 & 64.68 & 41.69 & 0.428 & 0.585 \\
    ConFEDE(U) & 81.1 & 80.95 & 68.93 & 45.43 & 0.387 & 0.643 \\
    Ours & \textbf{79.21} & \textbf{79.39} & \textbf{66.82} & \textbf{44.24} & \textbf{0.416} & \textbf{0.601} \\
    \hline
  \end{tabular}
  }
\end{table*}

\begin{figure*}[htbp]
  \centering
  \begin{minipage}[b]{0.45\textwidth}
    \includegraphics[width=\textwidth]{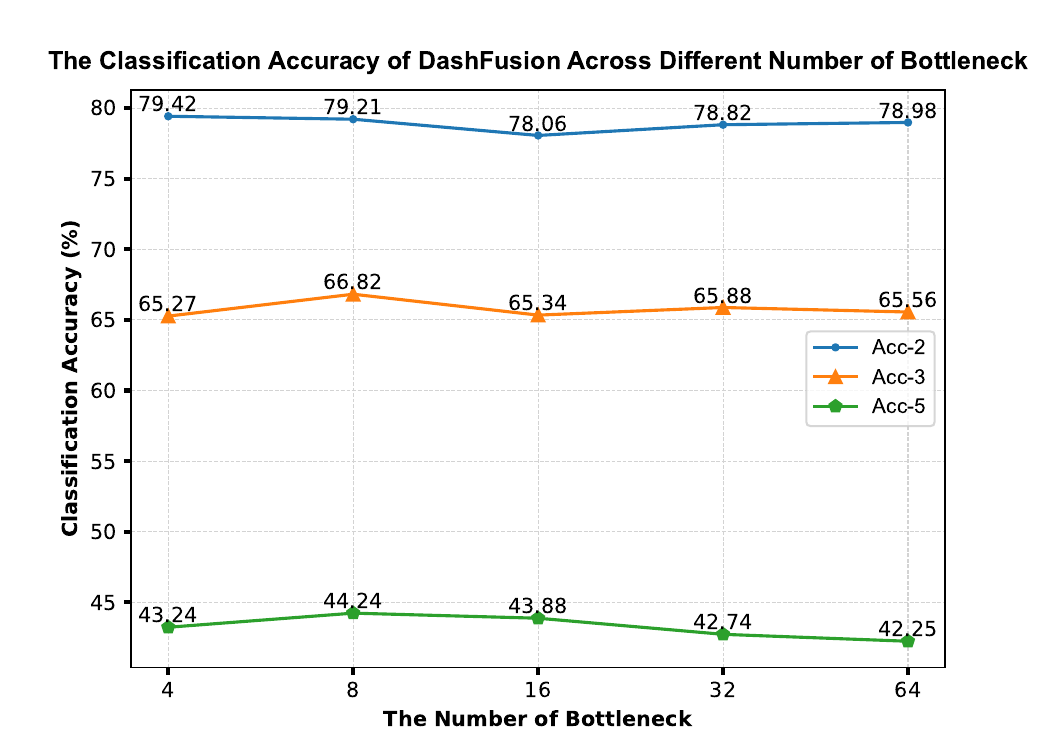}
  \end{minipage}
  \hspace{0.05\textwidth}
  \begin{minipage}[b]{0.45\textwidth}
    \includegraphics[width=\textwidth]{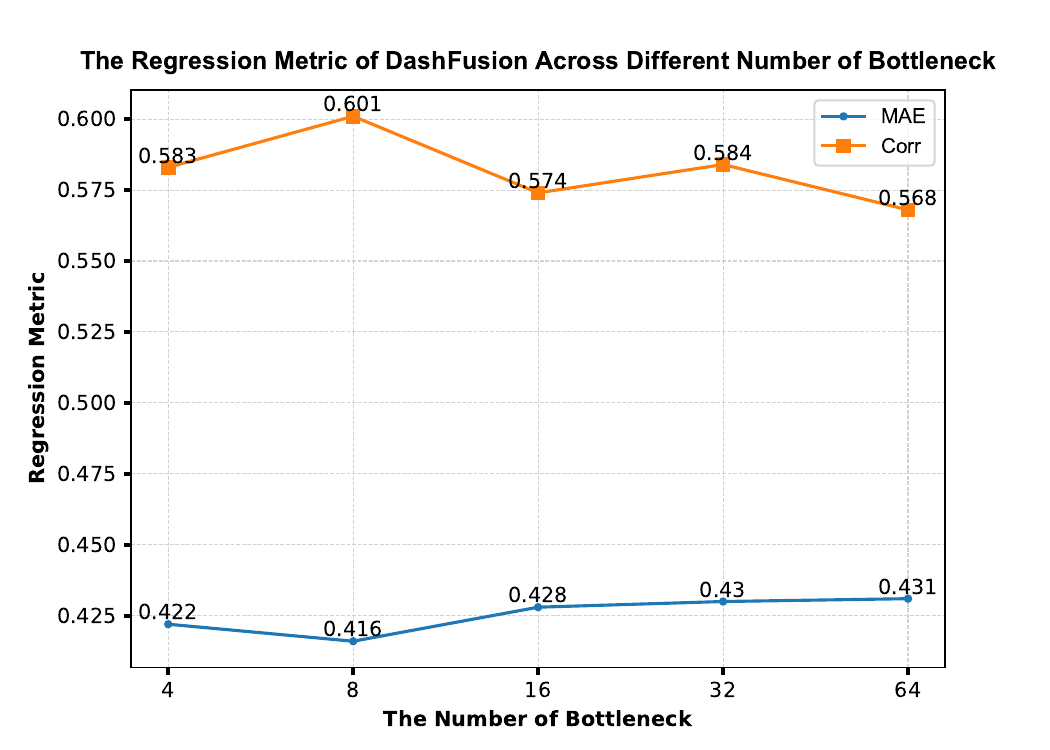}
  \end{minipage}
  \caption{Performance evaluation of DashFusion with varying numbers of bottleneck tokens on the CH-SIMS dataset. The left plot presents classification accuracy across different class settings, while the right plot illustrates regression metrics, including Mean Absolute Error (MAE) and Pearson correlation (Corr).}
  \label{fig:bottleneck}
\end{figure*}

\footnotetext[1]{\url{https://github.com/thuiar/MMSA/blob/master/results/result-stat.md}}

It is worth noting that Lian et al. \cite{lian2024merbench} found that the CMU-MOSI and CMU-MOSEI datasets strongly emphasize the text modality, limiting the contribution of multimodal fusion techniques. Since unimodal text models can achieve results comparable to multimodal models, additional modalities often provide limited benefits. In contrast, CH-SIMS exhibits a more distributed contribution across modalities, where audio and visual features carry more complementary sentiment cues, making unimodal performance significantly weaker. This characteristic necessitates effective multimodal fusion to achieve strong results. Therefore, CH-SIMS serves as a more suitable benchmark for evaluating our model’s ability to integrate multimodal information, and we select it for further ablation studies.

\subsection{Number of Bottleneck Tokens \( p \)}

To investigate the impact of the bottleneck count \( p \) on model performance, we conduct experiments with \( p \) = 4, 8, 16, 32, and 64. The results indicate that the evaluation metrics fluctuate slightly across different values of \( p \). As shown in Fig. \ref{fig:bottleneck}, setting \( p \) = 8 generally yields the best performance across most metrics. When \( p \) $<$ 8, performance improves as \( p \) increases. When \( p \) $>$ 8, performance slightly decreases. This phenomenon may be attributed to the introduction of redundant information, which interferes with sentiment assessment, especially in more nuanced multi-class classification and regression tasks. Therefore, we fix the number of tokens to \( p \) = 8 for subsequent experiments.

\subsection{Importance of Different Components}

To evaluate the effectiveness of each component in DashFusion, we conduct an ablation study by systematically removing key modules and analyzing their impact on performance. The results are shown in Table \ref{ablation1}. Each ablated variation results in a notable performance drop, highlighting the necessity of our proposed techniques.

Removing dual-stream alignment (the bottleneck is replaced by randomly initialized tokens) leads to a significant performance drop across all metrics, confirming that both temporal and semantic alignment play a crucial role in extracting consistent features and accurate sentiment analysis. Specifically, eliminating temporal alignment has a more conspicuous effect on multi-class accuracy, which needs subtle sentiment cues to predict. This finding suggests that temporal alignment of multimodal features enables the model to capture more precise and fine-grained sentiment information. On the other hand, omitting semantic alignment primarily affects two-class accuracy, implying that modality heterogeneity hinders sentiment prediction. By ensuring that features from different modalities share a coherent semantic space, semantic alignment enables better agreement between text, audio, and visual signals, especially in ambiguous cases.

Excluding supervised contrastive learning (SCL) results in a noticeable drop in performance, particularly in Acc-5. This suggests that SCL plays a crucial role in refining the model’s ability to differentiate sentiment classes, especially in complex multi-class settings. SCL enhances class separability by pulling samples with the same sentiment label closer in feature space while pushing apart samples from different classes. Additionally, due to the hard negative mining, it ensures that the model focuses on the most discriminative sentiment  features.

The absence of hierarchical bottleneck fusion (HBF) leads to the most significant performance decrease, confirming its critical function in efficiently integrating and compressing multimodal information. By progressively reducing the number of bottleneck tokens, HBF forces the model to filter out redundant information and retain only the most relevant sentiment information, thereby improving performance and generalization.

\begin{table}[!t]
  \centering
  \caption{The Ablation Study Results on CH-SIMS.\label{ablation1}}
  \resizebox{\columnwidth}{!}{
  \begin{tabular}{lcccc}
    \hline
    Method & F1($\uparrow$) &  Acc-5($\uparrow$) & MAE($\downarrow$) \\
    \hline
    DashFusion & \textbf{79.39}  & \textbf{44.24} & \textbf{0.412} \\
    w/o dual-stream alignment & 76.37 & 42.01 & 0.436 \\
    w/o temporal alignment & 79.03 & 42.12 & 0.418 \\
    w/o semantic alignment & 77.99 & 43.11 & 0.420 \\
    w/o SCL & 78.65 & 41.79 & 0.424 \\
    w/o HBF & 77.76 & 42.67 & 0.431 \\
    \hline
  \end{tabular}
  }
\end{table}

\subsection{Effectiveness of Different Fusion Mechanisms}

To compare the effectiveness of different fusion mechanisms, we conduct experiments using various fusion mechanisms on the CH-SIMS dataset. The results, presented in Table \ref{ablation2}, show the following observations.

The simplest method, concatenation, achieves moderate performance, indicating that when unimodal features are well-learned, simply combining them can be effective. However, it is not the most optimal approach for integrating multimodal information. Notably, this method incurs no additional multiply-accumulate operations (MAdds).
Applying the self-attention (SA) mechanism to concatenated features significantly improves performance across all metrics, suggesting that self-attention enhances the learning of interactions among different modal features. However, this approach requires 324 million MAdds, indicating a substantial computational cost.
The cross-modal attention (CA) mechanism integrates unimodal features into multimodal features and uses them for prediction. Although this method has a relatively low computational cost of 73 million MAdds, it performs poorly in terms of Acc-5 and MAE. This suggests that directly using cross-modal attention might lead to a loss of some feature details.

Bottleneck fusion (BF), which removes the compression process from our hierarchical bottleneck fusion, shows slightly better performance than simple concatenation. This demonstrates that using a bottleneck for fusion can help integrate multimodal features to some extent. Our proposed hierarchical bottleneck fusion (HBF) method achieves great improvement across most metrics and MAdds compared to bottleneck fusion. It delivers the best results in Acc-5 and MAE, confirming that the hierarchical approach of progressively reducing bottleneck tokens and using bi-directional cross-modal attention is highly effective in integrating and compressing multimodal information. Notably, the computational cost for our HBF is 145 million MAdds, which is less than half of that required by the self-attention mechanism, demonstrating that HBF can achieve superior performance while maintaining computational efficiency.

\begin{table}[!t]
  \centering
  \caption{Effects of Different Fusion Mechanisms on CH-SIMS. The Computation Cost is Measured by Multiply-Add Operations (MAdds) with One Video as the Input. M Denotes Million.}
  \label{ablation2}
  \resizebox{\columnwidth}{!}{
  \begin{tabular}{lcccc}
    \hline
    Method  & F1($\uparrow$) &  Acc-5($\uparrow$) & MAE($\downarrow$) & MAdds \\
    \hline
    Concat & 77.46  & 42.67 & 0.43 & \textbf{0}\\
    Concat\&SA & \textbf{79.52} & 44.08 & 0.424 & 324M \\
    CA & 78.53 & 39.95 & 0.456& 73M \\
    BF & 78.56 & 42.89 & 0.433& 162M \\
    HBF & 79.39 & \textbf{44.24} &  \textbf{0.412} & 145M\\
    \hline
  \end{tabular}
  }
\end{table}

\section{Conclusion}

In this paper, we present DashFusion, a novel framework designed to enhance multimodal sentiment analysis (MSA) through improved multimodal alignment and fusion. The dual-stream alignment module addresses temporal misalignment and modality heterogeneity across different modalities to achieve comprehension alignment. Supervised contrastive learning makes the modality features discriminative and robust. We design a progressive and efficient multimodal fusion method to integrate modality features through bottleneck and compress information layer by layer, which is called hierarchical bottleneck fusion. Extensive experiments on three benchmark datasets demonstrate the effectiveness of DashFusion, showing consistent performance improvements for both English and Chinese sentiment analysis. Ablation studies verify the contribution of each component, while comparative results show our hierarchical bottleneck fusion outperforms existing fusion methods. These findings collectively establish DashFusion as an effective solution for advancing MSA performance.

In future research, we will extend the applications of our proposed method. Besides MSA, we will leverage DashFusion for other types of sentiment understanding and affect computing tasks. Moreover, we also aim to adapt DashFusion to the missing modality setting and noisy modality setting, improving robustness in real-world applications where incomplete and noisy data are common challenges.

\bibliographystyle{IEEEtran}
\bibliography{custom}

@article{zadeh2016mosi,
  title={Multimodal sentiment intensity analysis in videos: Facial gestures and verbal messages},
  author={Zadeh, Amir and Zellers, Rowan and Pincus, Eli and Morency, Louis-Philippe},
  journal={IEEE Intell. Syst.},
  volume={31},
  number={6},
  pages={82--88},
  year={2016},
  publisher={IEEE}
}

@inproceedings{zadeh2018mosei,
    title = "Multimodal Language Analysis in the Wild: {CMU}-{MOSEI} Dataset and Interpretable Dynamic Fusion Graph",
    author = "Bagher Zadeh, AmirAli  and
      Liang, Paul Pu  and
      Poria, Soujanya  and
      Cambria, Erik  and
      Morency, Louis-Philippe",
    booktitle = "Proc. 56th Annu. Meeting Assoc. Comput. Linguistics",
    month = jul,
    year = "2018",
    pages = "2236--2246",
}

@inproceedings{yu2020ch-sims,
    title = "{CH}-{SIMS}: A {C}hinese Multimodal Sentiment Analysis Dataset with Fine-grained Annotation of Modality",
    author = "Yu, Wenmeng  and
      Xu, Hua  and
      Meng, Fanyang  and
      Zhu, Yilin  and
      Ma, Yixiao  and
      Wu, Jiele  and
      Zou, Jiyun  and
      Yang, Kaicheng",
    booktitle = "Proc. 58th Annu. Meeting Assoc. Comput. Linguistics",
    month = jul,
    year = "2020",
    pages = "3718--3727",
}

@inproceedings{chen2020simclr,
  title={A simple framework for contrastive learning of visual representations},
  author={Chen, Ting and Kornblith, Simon and Norouzi, Mohammad and Hinton, Geoffrey},
  booktitle={Proc. Int. Conf. Mach. Learn.},
  pages={1597--1607},
  year={2020},
  organization={PMLR}
}

@inproceedings{liu2018LMF,
    title = "Efficient Low-rank Multimodal Fusion With Modality-Specific Factors",
    author = "Liu, Zhun  and
      Shen, Ying  and
      Lakshminarasimhan, Varun Bharadhwaj  and
      Liang, Paul Pu  and
      Bagher Zadeh, AmirAli  and
      Morency, Louis-Philippe",
    booktitle = "Proc. 56th Annu. Meeting Assoc. Comput. Linguistics",
    month = jul,
    year = "2018",
    pages = "2247--2256",
}

@inproceedings{zadeh2017TFN,
    title = "Tensor Fusion Network for Multimodal Sentiment Analysis",
    author = "Zadeh, Amir  and
      Chen, Minghai  and
      Poria, Soujanya  and
      Cambria, Erik  and
      Morency, Louis-Philippe",
    booktitle = "Proc. 22th Conf. Empir. Methods Natural Lang. Process.",
    month = sep,
    year = "2017",
    pages = "1103--1114",
}

@inproceedings{tsai2019MuIT,
    title = "Multimodal Transformer for Unaligned Multimodal Language Sequences",
    author = "Tsai, Yao-Hung Hubert  and
      Bai, Shaojie  and
      Liang, Paul Pu  and
      Kolter, J. Zico  and
      Morency, Louis-Philippe  and
      Salakhutdinov, Ruslan",
    booktitle = "Proc. 57th Annu. Meeting Assoc. Comput. Linguistics",
    month = jul,
    year = "2019",
    pages = "6558--6569",
}

@inproceedings{hazarika2020MISA,
  title={Misa: Modality-invariant and-specific representations for multimodal sentiment analysis},
  author={Hazarika, Devamanyu and Zimmermann, Roger and Poria, Soujanya},
  booktitle={Proc. 28th ACM Int. Conf. Multimedia},
  pages={1122--1131},
  year={2020}
}

@inproceedings{rahman2020MAG-BERT,
    title = "Integrating Multimodal Information in Large Pretrained Transformers",
    author = "Rahman, Wasifur  and
      Hasan, Md Kamrul  and
      Lee, Sangwu  and
      Bagher Zadeh, AmirAli  and
      Mao, Chengfeng  and
      Morency, Louis-Philippe  and
      Hoque, Ehsan",
    booktitle = "Proc. 58th Annu. Meeting Assoc. Comput. Linguistic",
    month = jul,
    year = "2020",
    
    pages = "2359--2369",
}

@ARTICLE{mai2022HyCon,
  author={Mai, Sijie and Zeng, Ying and Zheng, Shuangjia and Hu, Haifeng},
  journal={IEEE Trans. Affective Comput.}, 
  title={Hybrid Contrastive Learning of Tri-Modal Representation for Multimodal Sentiment Analysis}, 
  year={2023},
  volume={14},
  number={3},
  pages={2276-2289},
  doi={10.1109/TAFFC.2022.3172360}}

@inproceedings{yu2021Self-MM,
  title={Learning modality-specific representations with self-supervised multi-task learning for multimodal sentiment analysis},
  author={Yu, Wenmeng and Xu, Hua and Yuan, Ziqi and Wu, Jiele},
  booktitle={Proc. AAAI Conf. Artif. Intell.},
  volume={35},
  number={12},
  pages={10790--10797},
  year={2021}
}

@inproceedings{yang2022FDMER,
  title={Disentangled representation learning for multimodal emotion recognition},
  author={Yang, Dingkang and Huang, Shuai and Kuang, Haopeng and Du, Yangtao and Zhang, Lihua},
  booktitle={Proc. 30th ACM Int. Conf. Multimedia},
  pages={1642--1651},
  year={2022}
}

@inproceedings{han2021MMIM,
    title = "Improving Multimodal Fusion with Hierarchical Mutual Information Maximization for Multimodal Sentiment Analysis",
    author = "Han, Wei  and
      Chen, Hui  and
      Poria, Soujanya",
    booktitle = "Proc. 2021 Conf. Empir. Methods Natural Lang. Process.",
    month = nov,
    year = "2021",
    pages = "9180--9192",
}

@inproceedings{yang2023ConFEDE,
    title = "{C}on{FEDE}: Contrastive Feature Decomposition for Multimodal Sentiment Analysis",
    author = "Yang, Jiuding  and
      Yu, Yakun  and
      Niu, Di  and
      Guo, Weidong  and
      Xu, Yu",
    booktitle = "Proc. 61st Annu. Meeting Assoc. Comput. Linguistic",
    month = jul,
    year = "2023",
    pages = "7617--7630",
}

@inproceedings{lv2021PMR,
  title={Progressive modality reinforcement for human multimodal emotion recognition from unaligned multimodal sequences},
  author={Lv, Fengmao and Chen, Xiang and Huang, Yanyong and Duan, Lixin and Lin, Guosheng},
  booktitle={Proc. IEEE/CVF Conf. Comput. Vis. Pattern Recognit.},
  pages={2554--2562},
  year={2021}
}

@inproceedings{guo2022CHFN,
  title={Dynamically adjust word representations using unaligned multimodal information},
  author={Guo, Jiwei and Tang, Jiajia and Dai, Weichen and Ding, Yu and Kong, Wanzeng},
  booktitle={Proc. 30th ACM Int. Conf. Multimedia},
  pages={3394--3402},
  year={2022}
}

@article{khosla2020supcon,
  title={Supervised contrastive learning},
  author={Khosla, Prannay and Teterwak, Piotr and Wang, Chen and Sarna, Aaron and Tian, Yonglong and Isola, Phillip and Maschinot, Aaron and Liu, Ce and Krishnan, Dilip},
  journal={Proc. Int. Conf. Neural Inf. Process. Syst.},
  volume={33},
  pages={18661--18673},
  year={2020}
}

@article{nagrani2021IBfusion,
  title={Attention bottlenecks for multimodal fusion},
  author={Nagrani, Arsha and Yang, Shan and Arnab, Anurag and Jansen, Aren and Schmid, Cordelia and Sun, Chen},
  journal={Proc. Int. Conf. Neural Inf. Process. Syst.},
  volume={34},
  pages={14200--14213},
  year={2021}
}

@article{shwartz2017opening,
  title={Opening the black box of deep neural networks via information},
  author={Shwartz-Ziv, Ravid and Tishby, Naftali},
  journal={arXiv preprint arXiv:1703.00810},
  year={2017}
}

@article{sun2023EMT,
  title={Efficient multimodal transformer with dual-level feature restoration for robust multimodal sentiment analysis},
  author={Sun, Licai and Lian, Zheng and Liu, Bin and Tao, Jianhua},
  journal={IEEE Trans. Affective Comput.},
  year={2023},
  publisher={IEEE}
}

@inproceedings{he2020MOCO,
  title={Momentum contrast for unsupervised visual representation learning},
  author={He, Kaiming and Fan, Haoqi and Wu, Yuxin and Xie, Saining and Girshick, Ross},
  booktitle={Proc. IEEE/CVF Conf. Comput. Vis. Pattern Recognit.},
  pages={9729--9738},
  year={2020}
}

@inproceedings{radford2021clip,
  title={Learning transferable visual models from natural language supervision},
  author={Radford, Alec and Kim, Jong Wook and Hallacy, Chris and Ramesh, Aditya and Goh, Gabriel and Agarwal, Sandhini and Sastry, Girish and Askell, Amanda and Mishkin, Pamela and Clark, Jack and others},
  booktitle={Proc. Int. Conf. Mach. Learn.},
  pages={8748--8763},
  year={2021},
  organization={PMLR}
}

@article{li2021albef,
  title={Align before fuse: Vision and language representation learning with momentum distillation},
  author={Li, Junnan and Selvaraju, Ramprasaath and Gotmare, Akhilesh and Joty, Shafiq and Xiong, Caiming and Hoi, Steven Chu Hong},
  journal={Proc. Int. Conf. Neural Inf. Process. Syst.},
  volume={34},
  pages={9694--9705},
  year={2021}
}

@inproceedings{zong2023acformer,
  title={AcFormer: An Aligned and Compact Transformer for Multimodal Sentiment Analysis},
  author={Zong, Daoming and Ding, Chaoyue and Li, Baoxiang and Li, Jiakui and Zheng, Ken and Zhou, Qunyan},
  booktitle={Proc. 31st ACM Int. Conf. Multimedia},
  pages={833--842},
  year={2023}
}

@article{liang2022modalitygap,
  title={Mind the gap: Understanding the modality gap in multi-modal contrastive representation learning},
  author={Liang, Victor Weixin and Zhang, Yuhui and Kwon, Yongchan and Yeung, Serena and Zou, James Y},
  journal={Proc. Int. Conf. Neural Inf. Process. Syst.},
  volume={35},
  pages={17612--17625},
  year={2022}
}

@article{lian2024merbench,
  title={Merbench: A unified evaluation benchmark for multimodal emotion recognition},
  author={Lian, Zheng and Sun, Licai and Ren, Yong and Gu, Hao and Sun, Haiyang and Chen, Lan and Liu, Bin and Tao, Jianhua},
  journal={arXiv preprint arXiv:2401.03429},
  year={2024}
}

@article{gandhi2023multimodal,
  title={Multimodal sentiment analysis: A systematic review of history, datasets, multimodal fusion methods, applications, challenges and future directions},
  author={Gandhi, Ankita and Adhvaryu, Kinjal and Poria, Soujanya and Cambria, Erik and Hussain, Amir},
  journal={Inf. Fusion},
  volume={91},
  pages={424--444},
  year={2023},
  publisher={Elsevier}
}

@inproceedings{mao2022m-sena,
    title = "{M}-{SENA}: An Integrated Platform for Multimodal Sentiment Analysis",
    author = "Mao, Huisheng  and
      Yuan, Ziqi  and
      Xu, Hua  and
      Yu, Wenmeng  and
      Liu, Yihe  and
      Gao, Kai",
    booktitle = "Proc. 60th Annu. Meeting Assoc. Comput. Linguistic: Syst. Demonstrations",
    month = may,
    year = "2022",
    pages = "204--213",
}

@ARTICLE{Lu2023SAsurvey,
  author={Lu, Qiang and Sun, Xia and Long, Yunfei and Gao, Zhizezhang and Feng, Jun and Sun, Tao},
  journal={IEEE Trans. Neural Netw. Learn, Syst.}, 
  title={Sentiment Analysis: Comprehensive Reviews, Recent Advances, and Open Challenges}, 
  year={2023},
  volume={},
  number={},
  pages={1-21},
  doi={10.1109/TNNLS.2023.3294810}}

@article{li2024enhancing,
  title={Enhancing Modal Fusion by Alignment and Label Matching for Multimodal Emotion Recognition},
  author={Li, Qifei and Gao, Yingming and Wen, Yuhua and Wang, Cong and Li, Ya},
  journal={arXiv preprint arXiv:2408.09438},
  year={2024}
}

@article{zhu2023multimodal,
  title={Multimodal sentiment analysis based on fusion methods: A survey},
  author={Zhu, Linan and Zhu, Zhechao and Zhang, Chenwei and Xu, Yifei and Kong, Xiangjie},
  journal={Inf. Fusion},
  volume={95},
  pages={306--325},
  year={2023},
  publisher={Elsevier}
}

@article{rosa2018knowledge,
  title={A knowledge-based recommendation system that includes sentiment analysis and deep learning},
  author={Rosa, Renata Lopes and Schwartz, Gisele Maria and Ruggiero, Wilson Vicente and Rodr{\'\i}guez, Dem{\'o}stenes Zegarra},
  journal={IEEE Trans. Ind. Inform.},
  volume={15},
  number={4},
  pages={2124--2135},
  year={2018},
  publisher={IEEE}
}

@article{ghosh2023multimodal,
  title={A multimodal sentiment analysis system for recognizing person aggressiveness in pain based on textual and visual information},
  author={Ghosh, Anay and Dhara, Bibhas Chandra and Pero, Chiara and Umer, Saiyed},
  journal={J. Ambient Intell. Humaniz. Comput.},
  volume={14},
  number={4},
  pages={4489--4501},
  year={2023},
  publisher={Springer}
}

@article{clavel2015sentiment,
  title={Sentiment analysis: from opinion mining to human-agent interaction},
  author={Clavel, Chloe and Callejas, Zoraida},
  journal={IEEE Trans. Affective Comput.},
  volume={7},
  number={1},
  pages={74--93},
  year={2015},
  publisher={IEEE}
}

@article{spatiotis2020sentiment,
  title={Sentiment analysis of teachers using social information in educational platform environments},
  author={Spatiotis, Nikolaos and Perikos, Isidoros and Mporas, Iosif and Paraskevas, Michael},
  journal={Int. J. Artif. Intell. Tools},
  volume={29},
  number={02},
  pages={2040004},
  year={2020},
  publisher={World Scientific}
}

@article{baltruvsaitis2018multimodal,
  title={Multimodal machine learning: A survey and taxonomy},
  author={Baltru{\v{s}}aitis, Tadas and Ahuja, Chaitanya and Morency, Louis-Philippe},
  journal={IEEE Trans. Pattern Anal. Mach. Intell.},
  volume={41},
  number={2},
  pages={423--443},
  year={2018},
  publisher={IEEE}
}

@article{zhao2020multi,
  title={Multi-source knowledge fusion: a survey},
  author={Zhao, Xiaojuan and Jia, Yan and Li, Aiping and Jiang, Rong and Song, Yichen},
  journal={World Wide Web},
  volume={23},
  pages={2567--2592},
  year={2020},
  publisher={Springer}
}

@article{liang2024foundations,
  title={Foundations \& trends in multimodal machine learning: Principles, challenges, and open questions},
  author={Liang, Paul Pu and Zadeh, Amir and Morency, Louis-Philippe},
  journal={ACM Comput. Surv.},
  volume={56},
  number={10},
  pages={1--42},
  year={2024},
  publisher={ACM New York, NY}
}

@article{zhu2024vision+,
  title={Vision+ x: A survey on multimodal learning in the light of data},
  author={Zhu, Ye and Wu, Yu and Sebe, Nicu and Yan, Yan},
  journal={IEEE Trans. Pattern Anal. Mach. Intell.},
  year={2024},
  publisher={IEEE}
}

@inproceedings{wu2018unsupervised,
  title={Unsupervised feature learning via non-parametric instance discrimination},
  author={Wu, Zhirong and Xiong, Yuanjun and Yu, Stella X and Lin, Dahua},
  booktitle={Proc. IEEE/CVF Conf. Comput. Vis. Pattern Recognit},
  pages={3733--3742},
  year={2018}
}

@INPROCEEDINGS{2014COVAREP,
  author={Degottex, Gilles and Kane, John and Drugman, Thomas and Raitio, Tuomo and Scherer, Stefan},
  booktitle={Proc. IEEE 40th Int. Conf. Acoust. Speech Signal Process.}, 
  title={COVAREP — A collaborative voice analysis repository for speech technologies}, 
  year={2014},
  volume={},
  number={},
  pages={960-964},
  doi={10.1109/ICASSP.2014.6853739}}

@INPROCEEDINGS{OpenFace,
  author={Baltrusaitis, Tadas and Zadeh, Amir and Lim, Yao Chong and Morency, Louis-Philippe},
  booktitle={Proc. IEEE 13th Int. Conf. Autom. Face Gesture Recognit.}, 
  title={OpenFace 2.0: Facial Behavior Analysis Toolkit}, 
  year={2018},
  volume={},
  number={},
  pages={59-66},
  doi={10.1109/FG.2018.00019}}

@inproceedings{mcfee2015librosa,
  title={librosa: Audio and Music Signal Analysis in Python},
  author={McFee, Brian and Raffel, Colin and Liang, Dawen and Ellis, Daniel and McVicar, Matt and Battenberg, Eric and Nieto, Oriol},
  booktitle={Proc. 14th Scipy. Conf.},
  pages={18},
  year={2015},
  organization={SciPy}
}

@article{oord2018representation,
  title={Representation learning with contrastive predictive coding},
  author={Oord, Aaron van den and Li, Yazhe and Vinyals, Oriol},
  journal={arXiv preprint arXiv:1807.03748},
  year={2018}
}

@article{ba2016layernorm,
  title={Layer normalization},
  author={Ba Jimmy Lei, Jamie Ryan Kiros and Geoffrey E Hinton},
  journal={arXiv preprint arXiv:1607.06450},
  year={2016}
}

@ARTICLE{Jin2023MSEN,
  author={Jin, Cong and Luo, Cong and Yan, Ming and Zhao, Guangzhe and Zhang, Guixuan and Zhang, Shuwu},
  journal={IEEE Trans. Neural Netw. Learn, Syst.}, 
  title={Weakening the Dominant Role of Text: CMOSI Dataset and Multimodal Semantic Enhancement Network}, 
  year={2023},
  volume={},
  number={},
  pages={1-15},
  doi={10.1109/TNNLS.2023.3282953}}

@article{kingma2014adam,
  title={Adam: A method for stochastic optimization},
  author={Kingma, Diederik P},
  journal={arXiv preprint arXiv:1412.6980},
  year={2014}
}

@ARTICLE{Cheng2024MMF,
  author={Cheng, Hongju and Yang, Zizhen and Zhang, Xiaoqi and Yang, Yang},
  journal={IEEE Trans. Affective Comput.}, 
  title={Multimodal Sentiment Analysis Based on Attentional Temporal Convolutional Network and Multi-Layer Feature Fusion}, 
  year={2023},
  volume={14},
  number={4},
  pages={3149-3163},
  doi={10.1109/TAFFC.2023.3265653}}

@ARTICLE{Wang2024CENet,
  author={Wang, Di and Liu, Shuai and Wang, Quan and Tian, Yumin and He, Lihuo and Gao, Xinbo},
  journal={IEEE Trans. Multimedia}, 
  title={Cross-Modal Enhancement Network for Multimodal Sentiment Analysis}, 
  year={2023},
  volume={25},
  number={},
  pages={4909-4921},
  doi={10.1109/TMM.2022.3183830}}

@inproceedings{hu-etal-2022-unimse,
    title = "{U}ni{MSE}: Towards Unified Multimodal Sentiment Analysis and Emotion Recognition",
    author = "Hu, Guimin  and
      Lin, Ting-En  and
      Zhao, Yi  and
      Lu, Guangming  and
      Wu, Yuchuan  and
      Li, Yongbin",
    booktitle = "Proc. 2022 Conf. Empir. Methods Natural Lang. Process.",
    month = dec,
    year = "2022",
    pages = "7837--7851",
}

@inproceedings{guo2022dynamically,
  title={Dynamically adjust word representations using unaligned multimodal information},
  author={Guo, Jiwei and Tang, Jiajia and Dai, Weichen and Ding, Yu and Kong, Wanzeng},
  booktitle={Proc. 30th ACM Int. Conf. Multimedia},
  pages={3394--3402},
  year={2022}
}

@inproceedings{he2016res,
  title={Deep residual learning for image recognition},
  author={He, Kaiming and Zhang, Xiangyu and Ren, Shaoqing and Sun, Jian},
  booktitle={Proc. IEEE/CVF Conf. Comput. Vis. Pattern Recognit},
  pages={770--778},
  year={2016}
}

@inproceedings{glorot2011deep,
  title={Deep sparse rectifier neural networks},
  author={Glorot, Xavier and Bordes, Antoine and Bengio, Yoshua},
  booktitle={Proc. 14th Int. Conf. Artif. Intell. Statis.},
  pages={315--323},
  year={2011},
  organization={JMLR Workshop and Conference Proceedings}
}

@inproceedings{li2023DMD,
  title={Decoupled multimodal distilling for emotion recognition},
  author={Li, Yong and Wang, Yuanzhi and Cui, Zhen},
  booktitle={Proc. IEEE/CVF Conf. Comput. Vis. Pattern Recognit},
  pages={6631--6640},
  year={2023}
}

@article{LIU2024TSA,
title = {Triadic temporal-semantic alignment for weakly-supervised video moment retrieval},
journal = {Pattern Recognit.},
volume = {156},
pages = {110819},
year = {2024},
issn = {0031-3203},
author = {Jin Liu and JiaLong Xie and Fengyu Zhou and Shengfeng He},
}

@inproceedings{kawaguchi2023IB,
  title={How does information bottleneck help deep learning?},
  author={Kawaguchi, Kenji and Deng, Zhun and Ji, Xu and Huang, Jiaoyang},
  booktitle={Proc. 40th Int. Conf. Mach. Learn.},
  pages={16049--16096},
  year={2023},
  organization={PMLR}
}

@ARTICLE{Hu2024IBsurvey,
  author={Hu, Shizhe and Lou, Zhengzheng and Yan, Xiaoqiang and Ye, Yangdong},
  journal={IEEE Trans. Pattern Anal. Mach. Intell.}, 
  title={A Survey on Information Bottleneck}, 
  year={2024},
  volume={46},
  number={8},
  pages={5325-5344},
  doi={10.1109/TPAMI.2024.3366349}
}

@article{mustafa2022MCL4MOE,
  title={Multimodal contrastive learning with limoe: the language-image mixture of experts},
  author={Mustafa, Basil and Riquelme, Carlos and Puigcerver, Joan and Jenatton, Rodolphe and Houlsby, Neil},
  journal={Proc. Int. Conf. Neural Inf. Process. Syst.},
  volume={35},
  pages={9564--9576},
  year={2022}
}

@article{sun2022MCL4video,
  title={Long-form video-language pre-training with multimodal temporal contrastive learning},
  author={Sun, Yuchong and Xue, Hongwei and Song, Ruihua and Liu, Bei and Yang, Huan and Fu, Jianlong},
  journal={Proc. Int. Conf. Neural Inf. Process. Syst.},
  volume={35},
  pages={38032--38045},
  year={2022}
}

@InProceedings{Jiang2024CL4MLLM,
    author    = {Jiang, Chaoya and Xu, Haiyang and Dong, Mengfan and Chen, Jiaxing and Ye, Wei and Yan, Ming and Ye, Qinghao and Zhang, Ji and Huang, Fei and Zhang, Shikun},
    title     = {Hallucination Augmented Contrastive Learning for Multimodal Large Language Model},
    booktitle = {Proc. IEEE/CVF Conf. Comput. Vis. Pattern Recognit},
    month     = {June},
    year      = {2024},
    pages     = {27036-27046}
}

@ARTICLE{Fan2024MCL,
  author={Fan, Cunhang and Zhu, Kang and Tao, Jianhua and Yi, Guofeng and Xue, Jun and Lv, Zhao},
  journal={IEEE Trans. Affective Comput.}, 
  title={Multi-Level Contrastive Learning: Hierarchical Alleviation of Heterogeneity in Multimodal Sentiment Analysis}, 
  year={2025},
  volume={16},
  number={1},
  pages={207-222},
  doi={10.1109/TAFFC.2024.3423671}
}

\vfill

\end{document}